\documentclass[journal,twoside,web]{ieeecolor}
\usepackage{tmi}
\usepackage{amssymb}
\usepackage{amsmath}
\usepackage{multirow}
\usepackage{booktabs}
\usepackage{array}
\usepackage{threeparttable}
\usepackage{graphicx}
\usepackage{graphics}
\usepackage[hidelinks]{hyperref}
\hypersetup{
    colorlinks=true,
    citecolor=green,
    filecolor=magenta
}
\usepackage{algorithm}
\usepackage{algpseudocode}
\usepackage{adjustbox}
\usepackage{pifont}
\usepackage{xcolor}

\begin{document}
\bibliographystyle{IEEEtran}

\title{TopoVST: Toward Topology-fidelitous Vessel Skeleton Tracking}
\author{Yaoyu Liu, Minghui Zhang, Junjun He, Yun Gu,~\IEEEmembership{Member,~IEEE}
\thanks{This work is supported by National Natural Science and Foundation of China (No.62373243).}
\thanks{Yaoyu Liu,  Minghui Zhang, and Yun Gu  are with Institute of Medical Robotics, Shanghai Jiao Tong University, Shanghai, China.(e-mail:\{lyyu19, minghuizhang, geron762\}@sjtu.edu.cn)} 
\thanks{Minghui Zhang, Yaoyu Liu and Yun Gu are also with the School of Automation and Intelligent Sensing, Shanghai Jiao Tong University, Shanghai, China. }
\thanks{Junjun He (hejunjun@pjlab.org) is with Shanghai AI Lab, Shanghai, China.}
\thanks{ Corresponding author: Yun Gu }
}

\maketitle

\begin{abstract}
    Automatic extraction of vessel skeletons is crucial for many clinical applications. However, achieving topologically faithful delineation of thin vessel skeletons remains highly challenging, primarily due to frequent discontinuities and the presence of spurious skeleton segments. To address these difficulties, we propose TopoVST, a topology-fidelitious vessel skeleton tracker. TopoVST constructs multi-scale sphere graphs to sample the input image and employs graph neural networks to jointly estimate tracking directions and vessel radii. The utilization of multi-scale representations is enhanced through a gating-based feature fusion mechanism, while the issue of class imbalance during training is mitigated by embedding a geometry-aware weighting scheme into the directional loss. In addition, we design a wave-propagation-based skeleton tracking algorithm that explicitly mitigates the generation of spurious skeletons through space-occupancy filtering. We evaluate TopoVST on two vessel datasets with different geometries. Extensive comparisons with state-of-the-art baselines demonstrate that TopoVST achieves competitive performance in both overlapping and topological metrics. Our source code is available at: \url{https://github.com/EndoluminalSurgicalVision-IMR/TopoVST}.
\end{abstract}


\section{Introduction}

The vessel skeleton plays an important role in characterizing the geometry of a patient's vascular system and serves as a key component in many clinical applications. In cardiovascular disease diagnosis, accurate modeling of coronary artery skeletons is an essential early step that underpins reliable and non-invasive clinical assessments~\cite{Raff2009Guide, Shahzad2013Stenoses, Kirili2013Standard}. 
In endovascular interventional robotic surgery, skeletons provide precise geometric references for both planning and intraoperative navigation~\cite{Li2024Robot, Song2024Pilot}.

Manual annotation of vessel skeletons is laborious and time-consuming; thus, automatic vessel skeleton extraction becomes necessary. With the development of deep learning-based medical analysis~\cite{Litjens2019SOTACardiac, Gharleghi2022SegReview, Alnasser2024SegReview}, automatic skeleton extraction directly from medical images with neural networks has emerged as a trend. From the perspective of the entire extraction pipeline, automatic vessel skeletonization methods can be divided into two categories. The first is segmentation-based skeleton extraction, which performs voxel-level segmentation~\cite{Isensee2020nnUNet,cicek20163dunet} followed by binary thinning~\cite{Lee1994Skel, Menten_2023_ICCV}. The second is tracking-based skeleton extraction, in which a tracker is initialized from seed points and iteratively predicts the skeletal directions~\cite{Wolterink2019CNNTracker, Alblas2025SIRE, Li2023COACT}.

\begin{figure}[t]
    \centering
    \includegraphics[width=1.0\linewidth]{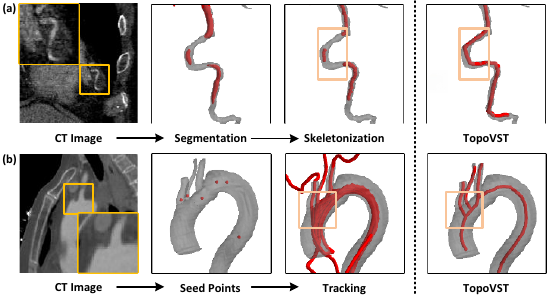}
    \caption{On the left, we provide a visual description and results of two vessel skeleton extraction pipelines, highlighting breakage and redundancy in colored boxes. On the right, we provide the results of TopoVST. (a)\textbf{Segmentation-based skeleton extraction}. The CT image is first segmented and then thinned. This method may cause $\beta_0$-errors due to false-negative breakage during segmentation. (b) \textbf{Tracking-based skeleton extraction}. A number of seed points are first acquired from the CT image. Then a tracker starts from these seed points and extracts the skeleton. This method is $\beta_1$-error-prone due to the aggregation of redundant skeleton branches during tracking.}
    \label{fig:challenges}
\end{figure}

\begin{figure*}[t]
\centering
\includegraphics[width=1.0\textwidth]{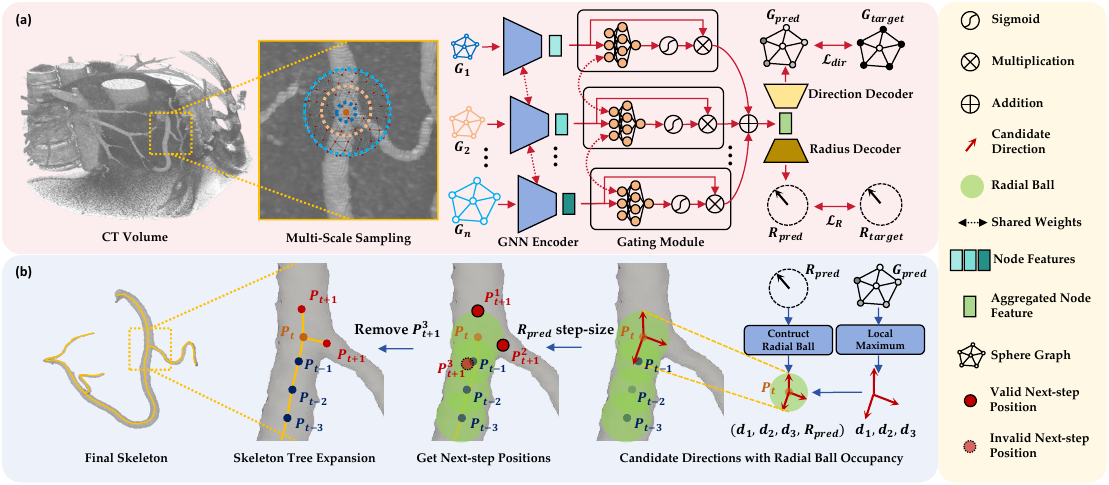}
\caption{\textbf{Our Framework}. (a) \textbf{Offline Training}. At a given position in the image, multi-scale sampling is performed to obtain graphs at different scales. These graphs are processed by a GNN encoder in parallel to extract node representations at different scales, which are passed through a gating module and summed to obtain aggregated node features. The aggregated node features are then used to decode node-level direction prediction $G_{pred}$ and graph-level radius prediction $R_{pred}$. (b) \textbf{Online Tracking}. We apply local maximum operation on the predicted graph to obtain candidate directions. The estimated radius provides both the tracking step size and radial-ball occupation in 3D space. Using the candidate directions and tracking step size, we determine the next-step positions in the vessel. These positions are subsequently filtered by the radial-ball occupations. We adopted a wave-propagation skeleton tree expansion algorithm to extract the final skeleton.}\label{fig:overall_pipeline}
\end{figure*}

The topology-fidelitous skeleton should faithfully capture the underlying vascular topology while preserving thinness and completeness~\cite{Tagliasacchi2016SkelReport}. However, existing automatic methods still struggle to meet these requirements.

\begin{itemize}
    \item \textbf{Breakage}. Fig.\ref{fig:challenges}(a) shows an example of breakage during segmentation-based skeleton extraction. The skeleton exhibits breakage due to segmentation errors in thin vessels, leading to incorrect topology. 
    Topology-aware losses~\cite{Shit_2021_CVPR, Shi2024cbDice, Acebes2024clCE, Kirchhoff2024SR, Zhang2024CAL} help reduce disconnections, but cannot fully prevent breakage, especially in thin or ambiguous regions. In tracking-based methods, breakage also arises when weak directional predictions at thin vessels or bifurcations cause the tracker to stop too early.

    \item \textbf{Redundant Skeletons}. Fig.\ref{fig:challenges}(b) shows an example of redundancy in tracking-based skeleton extraction. Without an explicit mechanism to distinguish explored from unexplored regions, the tracker may extract multiple skeletons for the same vessel branch. Such redundant skeletons introduce topological errors at bifurcations and violate the thinness property, thereby limiting their clinical utility. Existing approaches attempt to mitigate redundancy using hard-coded tracking constraints~\cite{Wolterink2019CNNTracker, Alblas2025SIRE} or post-processing heuristics~\cite{Liu2025DMAP}. However, hard-coded constraints and post-processing are often ad-hoc and not generalizable to vessels with varying radii.
\end{itemize}

To address the challenges described above, we propose TopoVST, a two-stage tracking-based vessel skeleton extraction framework, as shown in Fig.\ref{fig:overall_pipeline}. 

Breakage is mitigated during the offline training stage, where a graph neural network is trained to jointly predict the tracking directions and vessel radius.
We use gating-based multi-scale feature fusion in the network and geometry-aware weighting for the direction loss. The former re-weights and fuses multi-scale features for downstream direction and radius estimation. The latter alleviates class imbalance by constructing loss weights from the sphere geometry. In this way, we enhance the tracker's direction discrimination capability in thin vessels and preserve connectivity.

Redundancy is suppressed during the online tracking stage, where the tracker sequentially explores the vessel using predicted directions and radii. We expand a skeleton tree from seed points along the predicted directions, using estimated vessel radii as step sizes. We generate 3D ball-shaped occupancy regions from the predicted radii. These regions act as a filter to detect possible skeleton tree expansions toward explored regions inside the vessel. In this way, redundant skeletons can be detected and removed on-the-fly as the tracker explores inside the vessel.

Extensive experiments on two vessel datasets demonstrate that TopoVST achieves superior performance in overlapping metrics and $\beta_1$-errors, while maintaining comparable performance in $\beta_0$-errors.

We organize the remaining part of the article as follows: Section \ref{sec:related_works} briefly introduces related works on automatic skeleton extraction; Section \ref{sec:methods} introduces TopoVST from the offline training stage and the online tracking stage; Section \ref{sec:exp_res} and \ref{sec:results} provide detailed experimental settings and results analysis; Section \ref{sec:discussion} discusses the ablation results and elongation phenomenon; Section \ref{sec:conclusion} gives the final conclusion.


\section{Related Works}\label{sec:related_works}
\subsection{Segmentation-based Skeleton Extraction}

A widely adopted strategy for vascular skeleton extraction is to first generate vessel segmentation with neural networks\cite{Ronneberger2015UNet, cicek20163dunet, Isensee2020nnUNet} and then derive skeletons from them. As a result, the quality of the skeletons is highly dependent on the accuracy of the segmentation. To improve the topological fidelity in curvilinear structure segmentation, several methods introduce skeletons into the loss function to implicitly enforce vessel geometry. clDice~\cite{Shit_2021_CVPR} defines a novel Dice loss using predicted and target skeletons; SkeletonRecall~\cite{Kirchhoff2024SR} builds its loss upon the recall of target skeletons; CAL~\cite{Zhang2024CAL} constructs loss weights from target skeletons; cbDice\cite{Shi2024cbDice} further integrates the vessel radius to the clDice loss; clCE\cite{Acebes2024clCE} computes the cross-entropy loss over predicted and target skeletons. Despite various efforts in previous works toward better segmentation topology, they still fail to fully prevent breakage in the segmentation due to implicit geometry modeling. Recently, Vesselformer~\cite{Prabhakar2024Vesselformer} is proposed to directly predict skeleton graphs from images. However, it has to explicitly connect predicted graphs across different patches, and is thus still susceptible to breakage.

\subsection{Tracking-based Skeleton Extraction}
Another strategy for vessel skeleton extraction formulates the problem as guiding a tracker to sequentially explore the vessel. CNN Tracker~\cite{Wolterink2019CNNTracker} and a series of works~\cite{Liu2025DMAP, Gao2021Joint, Sun2023Livenet, Zhang2025_BACCE} use convolutional neural networks(CNN) to estimate skeleton directions and use these directions to guide the tracker. COACT~\cite{Li2023COACT} constructs a graph neural network(GNN) from a sphere mesh for direction estimation; SIRE~\cite{Alblas2025SIRE} extends a single-scale GNN to a multi-scale fashion, generalizing direction estimations to vessels with different radii. Recently, Trexplorer~\cite{Roman2024Trx} and Trexplorer-Super~\cite{Roman2025TrxSuper} sequentially generate the skeleton tree instead of predicting directions, but suffer from under-exploration of the vessel. Despite the continuity in the tracking process, breakage can still occur for methods using direction estimation since weak prediction of directions in indistinguishable vasculature prematurely terminates the tracker, preventing potential connections between explored skeleton segments. The redundancy problem is partially resolved in~\cite{Wolterink2019CNNTracker, Liu2025DMAP} with ad-hoc hard-coded constraints or post-processing techniques, but an online solution that is generalizable to different vessel geometries is still required.

\section{Methods}
\label{sec:methods}

\subsection{Overview}\label{subsec:overview_pipeline}

As shown in Fig.\ref{fig:overall_pipeline}, TopoVST can be divided into two stages: \textit{Offline Training} and \textit{Online Tracking}.
\begin{itemize}
    \item In offline training stage, TopoVST randomly samples from input image and trains the network to predict future directions and vessel radius. Multi-scale sampling is adopted, producing input sphere graphs at different scales: $G_1, G_2, \dots, G_n$. These graphs are subsequently processed by the GNN encoder in parallel. Next, gating modules are used to calculate the importance of node features at different scales before fusing them by weighted summation. These fused features are sent to downstream decoders for direction prediction $G_{pred}$ and radius prediction $R_{pred}$. The network is trained to minimize the differences between $G_{pred}$, $R_{pred}$ and $G_{target}$, $R_{target}$.
    \item In online tracking stage, TopoVST maintains a dynamic skeleton tree $T$, which expands along all verified skeleton directions in a wave-propagation manner. Estimated $G_{pred}$ is transformed into tracking directions by a local-maximum operation, and $R_{pred}$ is used as the step size. To prevent redundant detections, TopoVST utilizes previously estimated vessel radii to construct a space-occupancy filter and discards directions that may cause re-detection of the explored vessel. The remaining directions are then used to calculate the next-step positions and update $T$ with these positions. When the stopping criterion is satisfied, the wave-propagation procedure terminates and the expansion of $T$ ends.
\end{itemize}

\subsection{Offline Training}\label{subsec:offline}

\begin{figure}[t]
    \centering
    \includegraphics[width=1.0\linewidth]{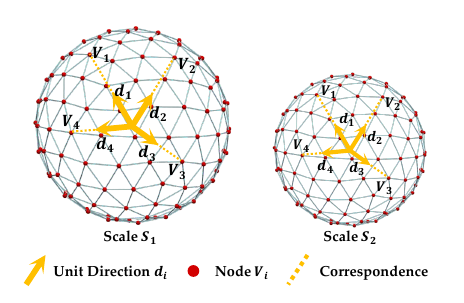}
    \caption{\textbf{Correspondence between unit directions and graph nodes} in two similar sphere graphs at different scales, $S_1$ and $S_2$. Four unit directions, $\mathbf{d_1}, \mathbf{d_2}, \mathbf{d_3}, \mathbf{d_4}$ are plotted with orange arrows. Four nodes, $V_1, V_2, V_3, V_4$ represent four points on the sphere mesh. Correspondence between each unit direction $\mathbf{d_i}$ and node $V_i$ can be found consistently in both scales, as denoted by orange dotted lines. Thus, direction modeling by the sphere graph is \textbf{scale-invariant}.}
    \label{fig:dir_node_correspondence}
\end{figure}

\begin{figure*}[t]
\centering
\includegraphics[width=0.9\textwidth]{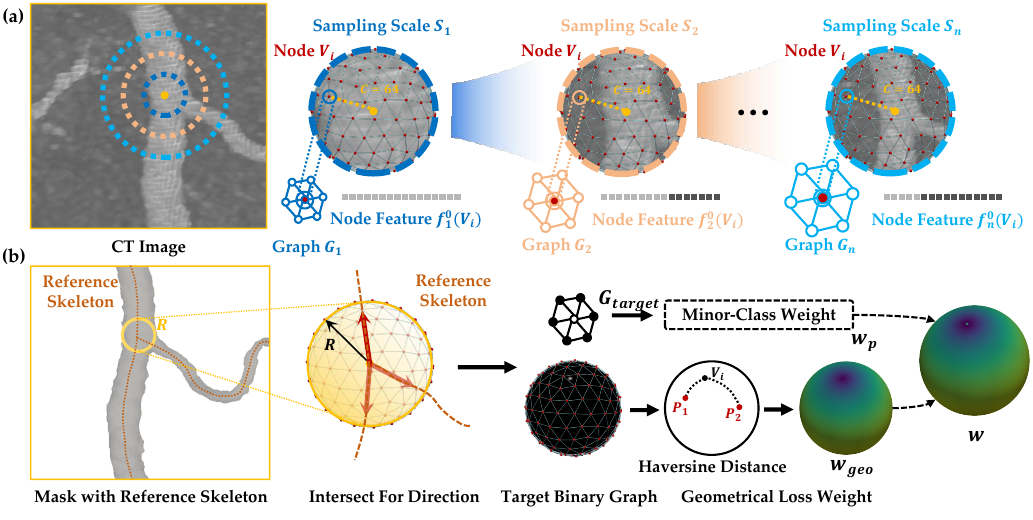}
\caption{\textbf{Illustration of multi-scale sampling, direction target formulation and loss weighting}. (a) \textbf{Multi-scale sampling}. Given a sampling position inside the vessel, $n$ sphere mesh graphs $G_1,G_2,\dots,G_n$ at scales $S_1, S_2, \dots, S_n$ are constructed. Each graph covers a different spherical region around the sampling point. For a node $V_i$ on graph $G_{m}$, we draw a line from the sampling position to $V_i$ and sample image intensities at $C=64$ points along the line. These intensities are concatenated into a feature vector in $\mathbb{R}^{C}$, denoted by $f_m^{0}(V_i)$. (b) \textbf{Direction target formulation and loss weighting}. For a given sampling position, we locate the intersections between its radial ball and the reference skeleton. Target directions can be calculated from these intersections and projected onto the predefined spherical graph. Nodes corresponding to the target directions are labeled as positive, and the remaining are labeled as negative. After projection, the target binary graph $G_{target}$ is acquired and can be used to compute the direction loss. Based on $G_{target}$, geometry-aware loss weights $w_{geo}$ can be computed from haversine distance. The loss weight $w$ can be obtained by combining $w_{geo}$ and the minority class weight $w_p$.}
\label{fig:sphere_sampling}
\end{figure*}

As shown in Fig.\ref{fig:dir_node_correspondence}, we build scale-invariant correspondences between 3D directions and sphere points using the sphere geometry. Organizing discrete sphere points into a graph paves the way for sampling and learning on geometrical structures with SO(3)-equivariance property~\cite{Alblas2025SIRE, Li2023COACT}. For better feature encoding, we use gauge-equivariant mesh convolution~\cite{cohen2019gauge, haan2021gauge} in the GNN encoder as follows:

\begin{equation}
    f^{k+1}_m(v_i) = \sigma\left(\sum_{v_j\in\mathcal{N}(v_i)\cup v_i} \Theta^{k+1} \rho(v_j\to v_i) f^{k}_m(v_i)\right)
\end{equation}

where $f^{k}_m(v_i)$ denotes the feature for node $v_i$ in graph $G_m$ in the $k$-th layer, $\Theta^{k}$ is the learnable kernel matrix in the $k$-th layer. $\rho(v_j\to v_i)$ represents a parallel transport of node features from $v_j$ to $v_i$ before aggregation. $\sigma$ is the non-linear function. We refer readers to~\cite{haan2021gauge} for more details.

\subsubsection{Gating-based Feature Fusion}

As shown in previous works~\cite{Salahuddin2021, Alblas2025SIRE}, multi-scale sampling enables generalizable direction estimation on different vessel geometries by providing the estimator with informative input. TopoVST follows~\cite{Alblas2025SIRE} to sample the input image using spheres with multiple radii centered at the sampling position. Fig.\ref{fig:sphere_sampling}(a) shows the process of sampling at scales $S_1, S_2, \dots, S_n$. The CT image at different scales represents different \textit{views} that the direction estimator can observe. The resulting node features $f_1^0(v_i), f_2^0(v_i), \dots, f_n^0(v_i)$ record different image intensity profiles that the estimator acquires when looking into the \textit{same} direction corresponding to $v_i$.

Unlike \cite{Alblas2025SIRE}, which predicts directions based on single-scale features and integrates them at the output, TopoVST performs multi-scale feature fusion within the network. To address the varying importance of input image views, we assign dynamic weights to features at different scales using gating, as shown in Fig.\ref{fig:overall_pipeline}. Within the gating module, the input node features at each scale are mapped to their corresponding weights in the subsequent feature fusion.

\begin{equation}
    f_{fused}(v_i) = \sum_{m=1}^{n} Sigmoid\left(Wf_m(v_i)\right) f_m(v_i)
\end{equation}
We omit superscripts here for simplicity. $f_m(v_i)$ denotes the feature for node $v_i$ in $G_m$ to be aggregated. $f_{fused}(v_i)$ denotes fused feature for node $v_i$, and $W$ is the weight of the linear projector.

\subsubsection{Geometry-aware Loss Weighting}

The left half of Fig.\ref{fig:sphere_sampling}(b) presents the process of formulating the direction target $G_{target}$. Future directions are acquired by intersecting the reference skeleton. Nodes on $G_{target}$ that are closest to future direction projections are marked as positive, while the remaining nodes are marked as negative. The difference between the number of positive- and negative-labeled nodes introduces \textit{class imbalance} into node classification, with positive labels as the minority. Furthermore, the distribution of positive-labeled nodes leads to \textit{spatial imbalance}, with loss signals on positive-labeled nodes smoothed by their surrounding negative-labeled neighbors.

In this paper, we introduced the weighting strategy to address imbalance issues. For class imbalance, $w_p > 1$ is applied only to positive-labeled nodes for emphasis. For spatial imbalance, $w_{geo}$ is applied to negative-labeled nodes that surround the positive-labeled nodes (see Eq.\eqref{eqn:wgeo}). Considering the spherical geometry, we use the haversine distance $\mathcal{H}_i$ to calculate the distance between a negative node $v_i$ and its nearest positive node in the graph.
\begin{equation}
    w_{geo}(v_i) = \left\{
    \begin{array}{ll}
        0 & \text{if } y_i = 1 \\
        \tanh(\mathcal{H}_i) & \text{if } y_i = 0
    \end{array}
    \right. \label{eqn:wgeo}
\end{equation}
where $y_i$ denotes the label for node $v_i$ in $G_{target}$. The final weight for direction loss can be formulated by combining $w_p$ and $w_{geo}$:
\begin{equation} \label{eqn:dir_loss_weight}
    w(v_i) = \left\{
    \begin{array}{ll}
        w_p & \text{if } y_i = 1 \\
        \tanh(\mathcal{H}_i) & \text{if } y_i = 0
    \end{array}
    \right.
\end{equation}
where $\mathcal{H}_i$ measures the distance between $v_i$ and its nearest positive-labeled node.

Since we estimate both tracking directions and vessel radius in offline training stage, the objective function for this stage can be formulated as a weighted sum of the direction estimation objective $\mathcal{L}_{dir}$ and the radius estimation objective $\mathcal{L}_R$. For $\mathcal{L}_{dir}$, we use the cross-entropy loss. For $\mathcal{L}_R$, we use the mean squared error loss. If we denote the predicted probability for node $v_i$ in $G_{pred}$ as $p_i$, then:

\begin{align}
    \mathcal{L}_{dir} &= -\frac{1}{N}\sum_{i\in N} w_i \left(y_i\cdot\log(p_i) + \left(1-y_i\right)\log\left(1-p_i\right)\right) \label{eqn:dir_loss} \\
    \mathcal{L}_R &= \frac{1}{2}\left(R_{target}-R_{pred}\right)^2 \\
    \mathcal{L} &= \lambda \mathcal{L}_{dir} + \mathcal{L}_R \label{eqn:total_loss}
\end{align}
We empirically set $w_p=10$ and $\lambda=5$ in our experiments.

\subsection{Online Tracking}\label{subsec:graph_sampling}

In online tracking stage, TopoVST adopts a breadth-first strategy to expand the skeleton tree $T$ from seed points. Fig.\ref{fig:waveprop} provides a visual example of the tracking process. In each iteration, TopoVST utilizes all detected directions at the current location and expands $T$ to all valid next-iteration positions. Leaf nodes in $T$ represent sampling positions for the next iteration. We view the expansion of $T$ as analogous to the propagation of a wave from a source, and term the algorithm \textit{wave-propagation}. The leaf nodes in $T$ are termed \textit{wave front}s.

\subsubsection{Space-occupancy Filter}
If the spatial extent of the previously explored vessel can be explicitly characterized, redundant skeletons can be effectively mitigated. A skeleton point is associated with a corresponding vessel radius, which naturally defines a ball-like region in 3D space, as shown in Fig.\ref{fig:waveprop}. By aggregating these regions along $T$, an approximate spatial representation of the previously explored vessel can be constructed. 
For a tracker at position $p_t$ with predicted vessel radius $R_t$, its previous positions and predicted radii are stored in $T$ as $p_0, p_1, \dots, p_{t-1}$ and $R_0, R_1, \dots, R_{t-1}$. Then a future direction $\mathbf{d}$ is valid if
\begin{align}
    &p_{t+1} = p_t + \mathbf{d}\cdot R_t \\
    &\lVert p_{t+1} - p_j \rVert \geq R_j,\quad j=0, 1, \dots, t-1 \label{eqn:space_filter}
\end{align}

We use this condition as a filter $\mathcal{F}_{s}$ to remove wave fronts $p_{t+1}$ that revisit explored regions within the vessel. Dotted arrows in Fig.\ref{fig:waveprop} illustrate invalid directions that create invalid wave fronts. In this way, $T$ is expanded without redundancy.

\subsubsection{Wave-propagation Algorithm}
\begin{figure}[t]
    \centering
    \includegraphics[width=1.0\linewidth]{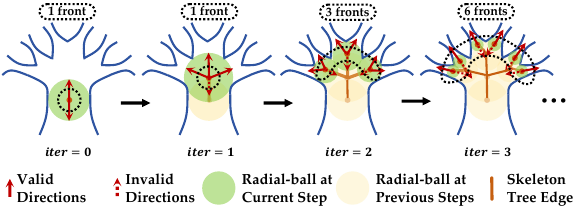}
    \caption{\textbf{Illustration of Wave-propagation Skeleton Tracking Algorithm}. For simplicity, downward tracking is not demonstrated. From left to right, we show the wave-propagation tracking procedure in each iteration. Wave fronts are enclosed in black dotted lines. Radial-ball occupancy is used to remove invalid tracking directions (red dotted arrows) that lead the tracker back to explored regions. The tracked skeleton tree is updated in every iteration.}
    \label{fig:waveprop}
\end{figure}

The wave-propagation algorithm is briefly described in Algorithm.\ref{alg:wave}. For a wave front $p_t$ in iteration $t$, we sample the image $I$ to obtain input $G_1, G_2,\dots,G_n$. The estimator $g$ then outputs the direction prediction $G_{pred}$ and the radius prediction $R_{pred}$. Local maximum operation is performed on the thresholded $G_{pred}$ to find nodes with locally maximal probability. These nodes are chosen as candidates for future directions: $\mathbf{d_1}, \mathbf{d_2}, \dots, \mathbf{d_k}$, where $k$ is the number of local maxima. If $k > 2$, a possible bifurcation is reached (see $iter$=1 and $iter$=2 in Fig.\ref{fig:waveprop}). If $k=0$, the tracker terminates its exploration at $p_t$ and switches to other wave fronts. Next, we use $R_{pred}$ as the step size and compute candidate wave fronts $p_{t+1}^{1}, p_{t+1}^{2}, \dots, p_{t+1}^k$. $\mathcal{F}_s$ is applied to these candidates, and wave fronts that do not satisfy Eq.\eqref{eqn:space_filter} are removed.

In each iteration, the tracker scans the current wave front list $L$ and updates both $T$ and $L$. Once there are no valid wave fronts after filtering with $\mathcal{F}_{s}$, the skeleton tracking process ends. Taking into account the computational burden, we manually set an upper limit on the length of $L$ as $L_{max}=20$.

\begin{algorithm}[!t]
    \caption{Wave-propagation Skeleton Tree Expansion Algorithm}\label{alg:wave}
    \begin{algorithmic}
        \Require Starting position $p_0$. Input image $I$. Estimator $g$. Maximum wave front length $L_{max}$
        \Ensure Extraction of skeleton tree $T$
        \State Initialize wave-front list $L=[p_0]$ and empty $T$
        \State $t \gets 0$
        \While {$0 < len(L) \leq L_{max}$}
            \State Initialize empty next-step wave front list $L^{\prime}=[]$.
            \For{$p_t$ in $L$}
                \State $G_1, G_2,\dots,G_n \gets $ Sample $I$ at $p_t$
                \State $\mathbf{d_1}, \mathbf{d_2}, \dots, \mathbf{d_k}, R_{t} \gets g(G_1, G_2,\dots,G_n)$
                \State $\left\{p_{t+1}\right\} = \left\{p_t + \mathbf{d} \cdot R_{t} | \mathbf{d}\in(\mathbf{d_1}, \mathbf{d_2}, \dots, \mathbf{d_k})\right\}$
                \State Add $\left\{p_{t+1}\right\}$ to $L^{\prime}$, $(p_t, R_t)$ to $T$
            \EndFor
            \State $L \gets \mathcal{F}_{s}\left(L^{\prime}\right)$
            \State $t \gets t+1$
        \EndWhile
    \end{algorithmic}
\end{algorithm}

\makeatletter
\def\hlinew#1{%
	\noalign{\ifnum0=`}\fi\hrule \@height #1 \futurelet
	\reserved@a\@xhline}
\makeatother

\begin{table*}[!t]
\centering
\caption{Quantitative Results (\textbf{Bold} represents $1$st place.)     OV: Overlap. Prec: Precision. $\beta_0$-err: $\beta_0$-error. $\beta_1$-err: $\beta_1$-error. Standard deviations are enclosed in brackets.  For each method, Avg.Rank calculates the average of its ranking position among all metrics.}\label{tab:full_res}
\renewcommand\arraystretch{1.5}

\begin{adjustbox}{max width=0.9\textwidth}
\begin{tabular}{>{\centering\arraybackslash}p{0.5cm}>{\raggedright\arraybackslash}p{2.5cm}>{\centering\arraybackslash}p{2.0cm}>{\centering\arraybackslash}p{2.0cm}>{\centering\arraybackslash}p{2.2cm}>{\centering\arraybackslash}p{2.2cm}>{\centering\arraybackslash}p{1.4cm}}
\hlinew{1pt}
Data & Methods & OV $\uparrow$ & Prec $\uparrow$ & $\beta_0$-err $\downarrow$ & $\beta_1$-err $\downarrow$ & Avg.Rank $\downarrow$ \\
\hlinew{1pt}
\multirow{8}{*}{\rotatebox[origin=c]{90}{ASOCA}}
 & VesselFM\cite{Wittmann2025vesselFM} & 0.8013($\pm 0.06$) & \textbf{0.8936}($\pm 0.04$) & 10.1250($\pm 4.97$) & 0.1250($\pm 0.35$) & 3.0 \\
 & CAL\cite{Zhang2024CAL} & 0.8424($\pm 0.05$) & 0.7629($\pm 0.09$) & 25.3750($\pm 9.18$) & 0.5000($\pm 0.76$) & 3.8 \\
 & SkeletonRecall\cite{Kirchhoff2024SR} & 0.8696($\pm 0.05$) & 0.6940($\pm 0.08$) & 24.7500($\pm 9.32$) & 0.7500($\pm 0.89$) & 4.5 \\
 & SIRE\cite{Alblas2025SIRE} & 0.8007($\pm 0.03$) & 0.6997($\pm 0.06$) & 3.8750($\pm 2.42$) & 209.0000($\pm 61.25$) & 5.0 \\
 & CNN Tracker\cite{Wolterink2019CNNTracker} & 0.3378($\pm 0.04$) & 0.2215($\pm 0.03$) & 2.0000($\pm 1.22$) & 46.0000($\pm 13.85$) & 5.8 \\
 & COACT Tracker\cite{Li2023COACT} & 0.7739($\pm 0.09$) & 0.7259($\pm 0.12$) & 389.3750($\pm 111.72$) & \textbf{0.0000}($\pm 0.00$) & 4.8 \\
 & Trexplorer-Super\cite{Roman2025TrxSuper} & 0.4431($\pm 0.13$) & 0.4027($\pm 0.19$) & \textbf{0.5000}($\pm 0.05$) & 0.7500($\pm 1.09$) & 4.8 \\
 \cline{2-7}
 & Ours & \textbf{0.8746}($\pm 0.03$) & 0.7404($\pm 0.09$) & 2.6250($\pm 2.78$) & \textbf{0.0000}($\pm 0.00$) & \textbf{2.0} \\
 \hlinew{1pt}
 \multirow{9}{*}{\rotatebox[origin=c]{90}{AortaSeg24}}
 & MedSAMv2\cite{Ma2025MedSAMv2} & 0.2391($\pm 0.21$) & 0.2862($\pm 0.31$) & \textbf{0.0000}($\pm 0.00$) & 24.6000($\pm 30.66$) & 6.0 \\
 & VesselFM\cite{Wittmann2025vesselFM} & 0.8239($\pm 0.10$) & 0.9418($\pm 0.03$) & 15.1000($\pm 7.84$) & 9.5000($\pm 7.68$) & 3.8 \\
 & CAL\cite{Zhang2024CAL} & 0.7880($\pm 0.08$) & 0.7615($\pm 0.12$) & 13.0000($\pm 7.06$) & 4.7000($\pm 3.95$) & 3.8 \\
 & SkeletonRecall\cite{Kirchhoff2024SR} & 0.8269($\pm 0.06$) & 0.6885($\pm 0.11$) & 21.9000($\pm 5.78$) & 1.5000($\pm 1.43$) & 3.8 \\
 & SIRE\cite{Alblas2025SIRE} & 0.6586($\pm 0.13$) & 0.6113($\pm 0.14$) & 26.9000($\pm 54.60$) & 45.6000($\pm 13.59$) & 6.0 \\
 & CNN Tracker\cite{Wolterink2019CNNTracker} & 0.4198($\pm 0.05$) & 0.3827($\pm 0.05$) & 167.4000($\pm 23.64$) & 1.0440($\pm 0.5422$) & 6.0 \\
 & COACT Tracker\cite{Li2023COACT} & 0.4855($\pm 0.03$) & 0.4068($\pm 0.04$) & 305.0000($\pm 24.10$) & \textbf{0.0000}($\pm 0.00$) & 5.3 \\
 & Trexplorer-Super\cite{Roman2025TrxSuper} & 0.1643($\pm 0.14$) & 0.1906($\pm 0.17$) & \textbf{0.0000}($\pm 0.00$) & 0.3000($\pm 0.46$) & 5.3 \\
 \cline{2-7}
 & Ours & \textbf{0.9469}($\pm 0.02$) & \textbf{0.9535}($\pm 0.02$) & 1.1000($\pm 1.14$) & \textbf{0.0000}($\pm 0.00$) & \textbf{1.3} \\
 \hlinew{1pt}
\end{tabular}
\end{adjustbox}

\begin{minipage}{0.95\textwidth}
    ~\\
    $\ast$ MedSAMv2 failed to segment coronary arteries on ASOCA, thus we do not report its metrics.
\end{minipage}
\end{table*}


\section{Experiments Settings}\label{sec:exp_res}

\subsection{Data}\label{subsec:data}

We incorporate two different vessel datasets in this work: ASOCA\cite{Gharleghi2023ASOCA} and AortaSeg24\cite{Muhammad2025Aortaseg24}, focusing on the coronary arteries and the aorta, respectively. These two types of vessels are characterized by different vessel radius distributions. The vessel radii in the AortaSeg24 dataset cover a wide range from $0$ mm to $25$ mm, while in the ASOCA dataset, the vessel radii fall between $0$ mm and $5$ mm.

\textbf{ASOCA}~\cite{Gharleghi2023ASOCA}: ASOCA is a dataset of computed tomography coronary arteries. It contains 40 cases with public annotations, among which 20 are diseased cases with evidence of stenosis or calcification and 20 are normal cases with no evidence of stenosis or obstructive disease. Vessels in this dataset manifest narrow lumens and tortuosity in certain regions.
We randomly select 16 normal cases and 16 diseased cases as the training set and the rest 8 cases as testing set. The annotated skeletons in the original dataset are used as reference skeletons for both training and evaluation.

\textbf{AortaSeg24}~\cite{Muhammad2025Aortaseg24} :
AortaSeg24 is a multi-class dataset of aortic vessels. It contains 50 public CTA image volumes with annotations of 23 anatomical regions. Vessels in this dataset are characterized by large radius contrasts between conjoined branches, as well as adjacency of thin branches. We randomly split the dataset into 40 training cases and 10 testing cases. Since no skeleton is available in the original dataset, we transform the annotations to single-class and use VMTK~\cite{Izzo2018VMTK} to extract reference skeleton models with 3D Slicer\footnote{\url{https://www.slicer.org/}}.

\subsection{Baselines}\label{subsec:baselines}

For segmentation-based methods, we use two different types of model. For task-specific vessel segmentation with topology-aware losses, we use 3dUNet~\cite{cicek20163dunet} paired with CAL~\cite{Zhang2024CAL} and SkeletonRecall~\cite{Kirchhoff2024SR}. For foundation segmentation models, we use MedSAMv2~\cite{Ma2025MedSAMv2} and VesselFM~\cite{Wittmann2025vesselFM}. MedSAMv2 focuses on general medical image segmentation, while VesselFM focuses on vascular structures. For thinning the segmentation results, we choose Lee's method~\cite{Lee1994Skel}, which is a widely adopted method for extracting skeletons from binary images.

For tracking-based methods, we choose CNN Tracker~\cite{Wolterink2019CNNTracker} and Trexplorer-Super~\cite{Roman2025TrxSuper} as baselines that use cubical patches for sampling. 
For baselines that use sphere graphs, we choose SIRE~\cite{Alblas2025SIRE} and COACT~\cite{Li2023COACT}. Since SIRE extracts a bidirectional single-branch skeleton for a given seed point, a post-processing step is adopted to connect these single-branch skeletons.

\begin{figure*}[!t]
    \centering
    \includegraphics[width=1.0\linewidth]{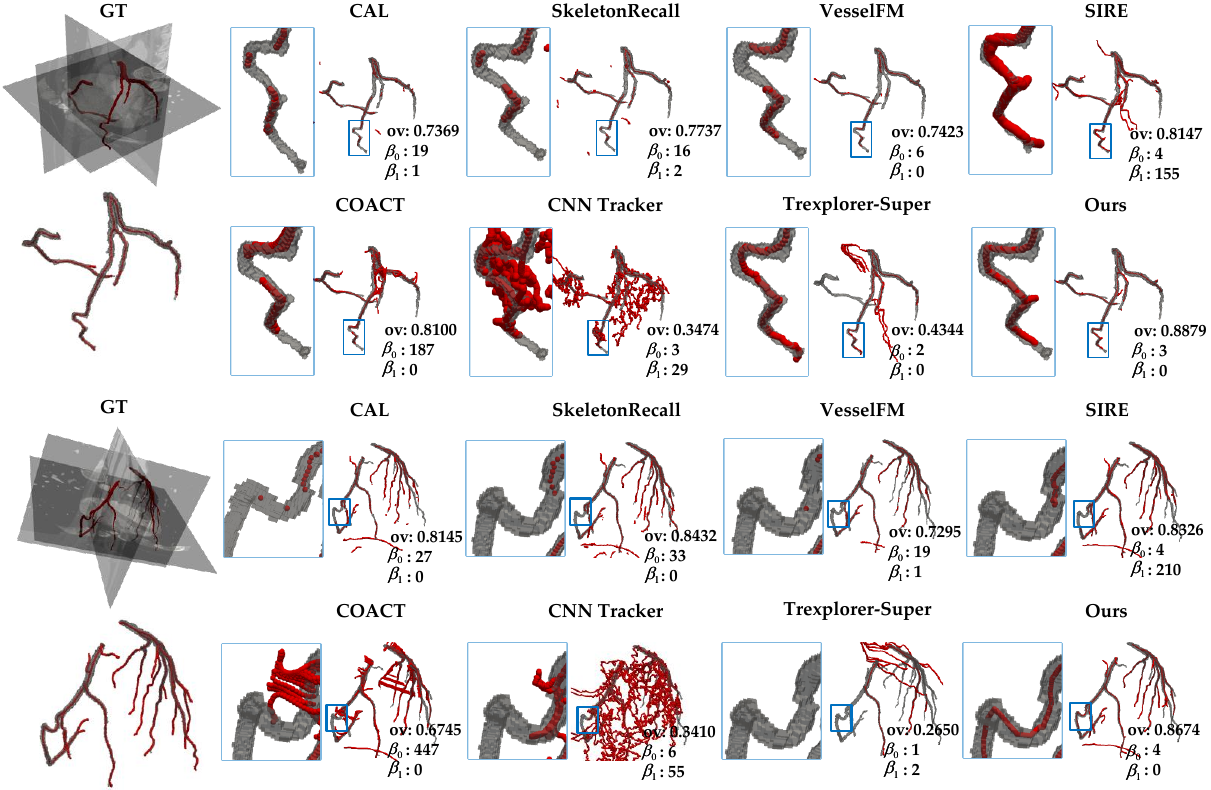}
    \caption{\textbf{Qualitative Evaluation on ASOCA Dataset}. The first case demonstrates a thin vessel segment in the coronary arteries. The second case demonstrates a tortuous vessel segment in the coronary arteries. TopoVST successfully extracts skeletons in these vessel segments. ov: Overlap.}
    \label{fig:main_res_ASOCA}
\end{figure*}

\begin{figure}[!t]
    \centering
    \includegraphics[width=1.1\linewidth]{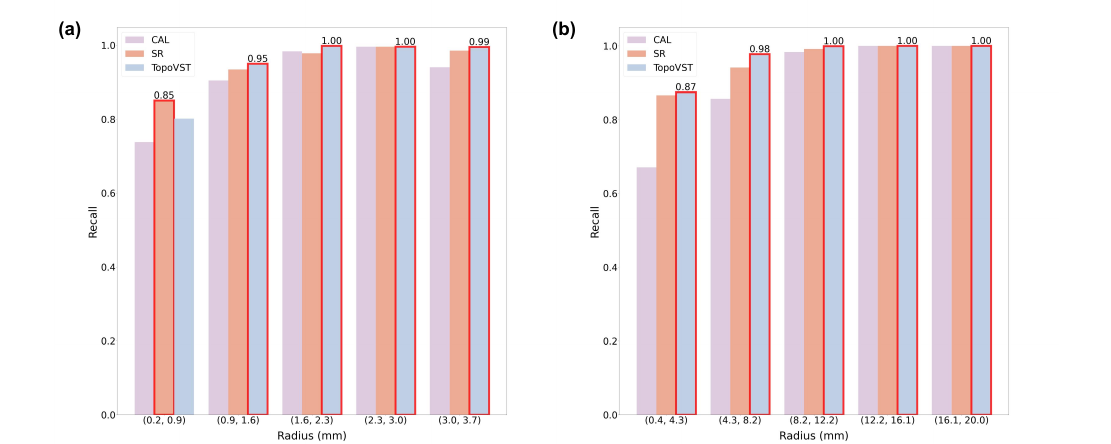}
    \caption{\textbf{Recall of the Reference Skeleton}. Here, we provide the recall of the reference skeleton. For effective comparison, we split the reference skeleton into five groups based on radius distributions and derive the statistics. (a) show the recall on ASOCA, and (b) shows the recall on AortaSeg24. It can be observed that TopoVST achieves the highest recall in most of the groups.}
    \label{fig:recall_compare}
\end{figure}

\begin{figure*}[!t]
    \centering
    \includegraphics[width=1.0\linewidth]{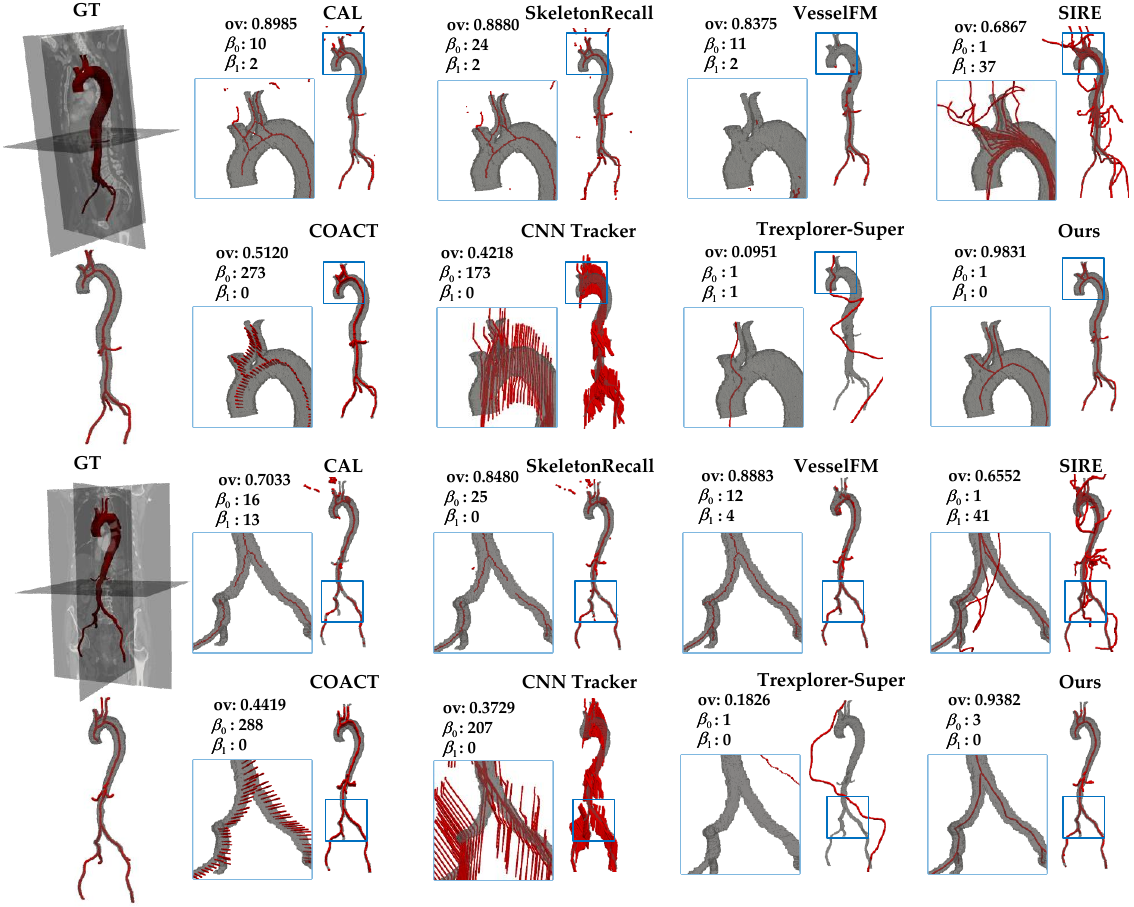}
    \caption{\textbf{Qualitative Evaluation on AortaSeg24 Dataset}. The first case demonstrates spatially adjacent vessel segments in the aorta. The second case demonstrates calcified vessel segments in the aorta. TopoVST extracts accurate skeletons in these vessel segments without redundancy. ov: Overlap.}
    \label{fig:main_res_Aorta24}
\end{figure*}

\subsection{Implementation Details}

In offline training, positions on annotated skeletons are sampled with a uniform distribution. Thirty percent of the samples are shifted within the radial ball for off-skeleton augmentation, and ten percent are shifted outside the lumen. To enable robust direction and radius estimation, we randomly draw 3 to 15 sampling scales from a uniform distribution between $S_{min}$ and $S_{max}$, which are selected based on the radius distributions in different datasets.
For ASOCA, we set $S_{min}=0.2\text{ mm}, S_{max}=15.0\text{ mm}$. For AortaSeg24, we set $S_{min}=1.0\text{ mm}, S_{max}=30.0\text{ mm}$. To allow sufficient views, there should be at least one scale smaller than $R_{target}$ and one scale larger than $R_{target}$.

In online tracking, the sampling scales are set to [1, 2, 3, 4, 5, 6, 7, 8, 9, 10] for ASOCA, and [2.5, 3, 3.5, 4, 4.5, 5, 5.5, 6, 6.5, 7, 10, 15, 20, 25, 30] for AortaSeg24. Seed points are initialized from the thinning results of rough segmentation produced by nnUNet\cite{Isensee2020nnUNet}. For the AortaSeg24 dataset, we add a termination criterion based on the distance to the surface of the rough segmentation result and set the threshold to $1.0$ mm, since the tracker often tracks far beyond the boundary of the GT mask, as discussed in \ref{subsec:failed_case}. We implement our training and testing algorithms on two NVIDIA 3090 GPUs, with the batch size set to 8. The Adam optimizer~\cite{Kingma2014Adam} is used with an initial learning rate of $5\times10^{-4}$.

For evaluation, we follow~\cite{Schaap2009Metrics} and measure the overlap and precision of the extracted skeletons. In addition to these metrics, we also evaluate the $\beta_0$-error and $\beta_1$-error as topological metrics of the skeletons. In addition, we compute the average ranking position for each metric to assess the overall performance of the skeleton.

\section{Results}\label{sec:results}

Quantitative results can be seen in Table \ref{tab:full_res}. Qualitative results for ASOCA and AortaSeg24 are presented in Fig.\ref{fig:main_res_ASOCA} and Fig.\ref{fig:main_res_Aorta24}, respectively. A general observation is that segmentation-based methods cannot achieve good geometrical accuracy. Specifically, foundation segmentation models lack perception of specific vessel structures, whereas task-specific segmentation models do not guarantee topological accuracy. Tracking-based methods without multi-scale perception perform poorly in overlap. In addition, tracking-based methods without proper vessel skeleton tracking algorithms perform poorly in topological accuracy.

\subsection{Compare with Segmentation-based Methods}

General-purpose foundation models do not guarantee performance on vascular structures, as is the case with MedSAMv2~\cite{Ma2025MedSAMv2}. MedSAMv2 fails to predict the vessels on ASOCA and achieves only $0.2391$ overlap on AortaSeg24. For this reason, we do not provide qualitative results in Fig.\ref{fig:main_res_ASOCA} and Fig.\ref{fig:main_res_Aorta24}. Vessel foundation models, represented by VesselFM~\cite{Wittmann2025vesselFM}, perform better than general-purpose models on vascular structures. VesselFM successfully captures vessel-like geometries, achieving an overlap of $0.8424$. However, it lacks knowledge of anatomy on specific vessels and segments all possible vascular regions. Hence, we crop VesselFM results based on nnUNet predictions before further evaluation. Furthermore, VesselFM does not incorporate many geometrical considerations into its framework, causing poor performance in both $\beta_0$-error and $\beta_1$-error.

Task-specific vessel segmentation models trained with topology-aware losses achieve better performance than foundation models on specific vessel datasets. Specifically, SkeletonRecall~\cite{Kirchhoff2024SR} achieves an overlap of $0.8696$ on ASOCA and $0.8269$ on AortaSeg24, surpassing both MedSAMv2~\cite{Ma2025MedSAMv2} and VesselFM~\cite{Wittmann2025vesselFM}. However, topology-aware losses alone do not guarantee topologically accurate segmentation results, as both CAL~\cite{Zhang2024CAL} and SkeletonRecall~\cite{Kirchhoff2024SR} perform poorly on $\beta_0$- and $\beta_1$-error. On ASOCA, the $\beta_0$- and $\beta_1$-errors of these two methods exceed most tracking methods. On AortaSeg24, their $\beta_0$- and $\beta_1$-errors are still much higher than those of TopoVST. Both Fig.\ref{fig:main_res_ASOCA} and Fig.\ref{fig:main_res_Aorta24} show the breakage in skeletons extracted by segmentation-based methods. Furthermore, Fig.\ref{fig:main_res_Aorta24} demonstrates in the top row how incorrect segmentation leads to $\beta_1$-errors. Unexpected cycles occur in regions where three thin aortic branches are adjacent.

TopoVST can extract more complete skeletons compared to task-specific vessel segmentation methods. Fig.\ref{fig:recall_compare} shows the recall of the reference skeleton under different methods. We compare TopoVST with CAL~\cite{Zhang2024CAL} and SkeletonRecall~\cite{Kirchhoff2024SR} on both ASOCA and AortaSeg24. For effective comparison, the reference skeletons in each dataset are divided into five groups based on the radius. The recall under TopoVST exceed those under CAL and SkeletonRecall in nine out of ten groups across the two datasets.

\subsection{Compare with Tracking-based Methods}

CNN Tracker~\cite{Wolterink2019CNNTracker}, COACT~\cite{Li2023COACT} and Trexplorer-Super~\cite{Roman2025TrxSuper} do not incorporate multi-scale perception, achieving overlap of $0.3378$, $0.7739$ and $0.4431$ on ASOCA, and $0.4198$, $0.4855$, $0.1643$ on AortaSeg24, respectively. The drop in overlap score mainly originates from low precision. In terms of precision, these methods predict skeletons with false positives, leading to lower precision and consequently lower overlap. Fig.\ref{fig:main_res_ASOCA} clearly shows that skeletons extracted by both CNN Tracker and COACT are prone to leakages. SIRE~\cite{Alblas2025SIRE} and TopoVST, equipped with multi-scale perception, achieve notably better overlap on both ASOCA and AortaSeg24.

The skeleton tracking algorithm also affects the topological quality of the extracted vessel skeletons. SIRE~\cite{Alblas2025SIRE} adopts single-branch tracking and radius-agnostic stopping conditions, leading to redundant skeleton detection. This is clearly demonstrated in Fig.\ref{fig:main_res_Aorta24} and Fig.\ref{fig:main_res_ASOCA}. When we connect the tracked skeleton branches into the final vessel skeleton tree, $\beta_1$-errors may occur. For COACT~\cite{Li2023COACT}, multi-branch detection is enabled in its skeleton tracking algorithm. However, radius-agnostic stopping conditions still lead to redundancy. The $\beta_1$-error of SIRE reaches $209.0000$ on ASOCA and $45.6000$ on AortaSeg24. For COACT, the $\beta_1$-error on both ASOCA and AortaSeg24 datasets is zero, but the $\beta_0$-error reaches $389.3750$ and $305.0000$, respectively. TopoVST, on the other hand, achieves zero $\beta_1$-error and relatively low $\beta_0$-error on both datasets. The $\beta_0$-error is $2.6520$ and $1.1000$ on ASOCA and AortaSeg24, significantly lower than that of segmentation-based methods and most tracking-based methods. 

\makeatletter
\def\hlinew#1{%
	\noalign{\ifnum0=`}\fi\hrule \@height #1 \futurelet
	\reserved@a\@xhline}
\makeatother
\begin{table}[!t]
\centering
\caption{Ablation Results on Gating and Loss Weighting}\label{tab:ablation_res}
\renewcommand\arraystretch{1.5}
\begin{adjustbox}{max width=0.95\linewidth}
\begin{tabular}{>{\centering\arraybackslash}p{0.5cm}>{\centering\arraybackslash}p{0.8cm}>{\centering\arraybackslash}p{0.8cm}>{\centering\arraybackslash}p{1.5cm}>{\centering\arraybackslash}p{1.5cm}>{\centering\arraybackslash}p{1.5cm}>{\centering\arraybackslash}p{1.5cm}}
\hlinew{1pt}
Data & Gating & Weighting & OV $\uparrow$ & Prec $\uparrow$ & $\beta_0$-err $\downarrow$ & $\beta_1$-err $\downarrow$ \\
\hlinew{1pt}
\multirow{4}{*}{\rotatebox[origin=c]{90}{ASOCA}}
 & \ding{55} & \ding{55} & 0.8707($\pm 0.03$)          & \textbf{0.8008}($\pm 0.08$) & 5.0000($\pm 3.54$)          & 0.0000($\pm 0.00$) \\
 & \ding{55} & \ding{51} & 0.8725($\pm 0.02$)          & 0.6671($\pm 0.08$)          & \textbf{2.1250}($\pm 2.47$) & 0.0000($\pm 0.00$) \\
 & \ding{51} & \ding{55} & 0.8710($\pm 0.04$)          & 0.7985($\pm 0.07$)          & 4.3750($\pm 3.71$)          & 0.0000($\pm 0.00$) \\
 & \ding{51} & \ding{51} & \textbf{0.8746}($\pm 0.03$) & 0.7404($\pm 0.09$)          & 2.6250($\pm 2.78$)          & 0.0000($\pm 0.00$) \\
 \hlinew{1pt}
 \multirow{4}{*}{\rotatebox[origin=c]{90}{AortaSeg24}}
 & \ding{55} & \ding{55} & 0.9368($\pm 0.03$)          & 0.9419($\pm 0.03$)          & 1.1000($\pm 0.83$)          & 0.0000($\pm 0.00$) \\
 & \ding{55} & \ding{51} & \textbf{0.9479}($\pm 0.02$) & 0.9542($\pm 0.02$)          & 1.2000($\pm 0.98$)          & 0.0000($\pm 0.00$) \\
 & \ding{51} & \ding{55} & 0.9462($\pm 0.02$)          & 0.9522($\pm 0.02$)          & 1.3000($\pm 1.10$)          & 0.0000($\pm 0.00$) \\
 & \ding{51} & \ding{51} & 0.9469($\pm 0.02$)          & \textbf{0.9535}($\pm 0.02$) & \textbf{1.1000}($\pm 1.14$) & 0.0000($\pm 0.00$) \\
 \hlinew{1pt}
\end{tabular}
\end{adjustbox}
\end{table}

\begin{figure}[!t]
    \centering
    \includegraphics[width=1.0\linewidth]{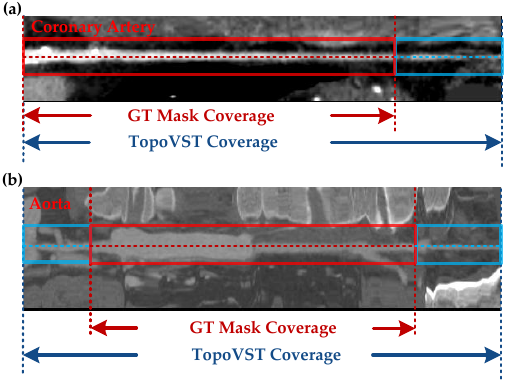}
    \caption{\textbf{Elongation of Tracked Vessel Skeleton}. Here we provide the Curved Planar Reformat(CPR) plot on two different datasets. Vessel regions labeled in GT and extracted by TopoVST are highlighted with red and blue rectangular boxes, respectively. (a) provides a view on ASOCA dataset and (b) provides a view on AortaSeg24 dataset. On these two plots it can be observed that skeleton extracted by our method extends beyond the coverage of GT mask, but still remains within the same part of vessels.}
    \label{fig:failed_case}
\end{figure}

\section{Discussion}\label{sec:discussion}
\subsection{Ablation Studies}\label{subsec:ablations}

We organize ablation experiments on two key components in TopoVST: gating-based feature fusion and direction loss weighting. We replace the former with a simple averaging operation and the latter with equal weights of value 1. Table \ref{tab:ablation_res} summarizes results of the ablation experiments. 
Regarding direction loss weighting, trackers with loss weighting generally perform better in overlap and $\beta_0$-error but worse in precision than trackers without it. This shows that adding loss weighting improves the direction estimation performance, encouraging the tracker to explore more vessel regions, and possibly causing leakage. Regarding gating-based feature fusion, trackers with gating-based feature fusion generally achieve slightly better overlap than trackers without it. If we compare the improvement on overlap between gating-based feature fusion and direction loss weighting, it is clear that loss weighting has a greater influence on skeleton quality.

\subsection{Failure Case Analysis}\label{subsec:failed_case}

A typical phenomenon during skeleton tracking is leakage outside the vessel lumen, which can result in lower precision when compared to GT skeletons generated from GT segmentation masks. In our experiments, leakage is also observed at the ends of the branches in labeled vessels, often extending further in the tracking direction before leakage occurs. This suggests that our tracker may continue to follow the same vessel, even if it is not labeled in the original dataset for some reason. To better illustrate this finding, we plot the Curved Planar Reformat(CPR) results along the tracked vessel branch. Fig.\ref{fig:failed_case} shows the CPR results on both ASOCA and AortaSeg24. We mark the coverage of GT segmentation mask on the vessel with red rectangular boxes and the extra coverage of the tracked skeleton on the vessel with blue rectangular boxes. From the CPR plots, it is clear that the skeleton tracked by TopoVST can extend into unlabeled regions of the same vessel, beyond the scope of GT segmentation. Thus, we term this phenomenon \textit{elongation} instead of leakage. Despite reducing precision and overlap, elongation helps physicians capture smaller and more complete vascular tree structures, which can be very useful for certain clinical applications.

\section{Conclusion}\label{sec:conclusion}

In this work, we propose TopoVST, a skeleton tracking framework that preserves vessel topology without redundancy. In offline training stage, we use a multi-scale sphere graph for direction and radius prediction and design gating-based feature fusion to efficiently utilize node features at different scales. Geometry-aware direction loss weighting is introduced to enable the network to learn effectively under class imbalance. In online tracking stage, a wave-propagation tracking algorithm with space-occupancy filtering is employed, which expands the skeleton tree in all valid directions. Empirical evaluations on two datasets demonstrate the effectiveness of our method in both overlap and topological metrics.

\section*{References}
\bibliography{ref.bib}
\end{document}